\documentclass[conference]{IEEEtran}
\IEEEoverridecommandlockouts

\usepackage{amsmath,amssymb,amsfonts}
\usepackage{algorithmic}
\usepackage{graphicx}
\usepackage{textcomp}
\usepackage{xcolor}
\usepackage{biblatex}
\usepackage{times}
\usepackage{latexsym}
\usepackage{multirow}
\usepackage{booktabs}

\usepackage{amsmath}
\usepackage{times}
\usepackage{latexsym}
\usepackage{times}
\usepackage{latexsym}
\usepackage{multirow}
\usepackage{booktabs}
\def\BibTeX{{\rm B\kern-.05em{\sc i\kern-.025em b}\kern-.08em
    T\kern-.1667em\lower.7ex\hbox{E}\kern-.125emX}}
\begin{document}

\title{M-Eval: A Heterogeneity-Based Framework for Multi-evidence Validation in Medical RAG Systems}

\author{\IEEEauthorblockN{1\textsuperscript{st}Mengzhou Sun}
\IEEEauthorblockA{\textit{Faculty of Computing} \\
\textit{Harbin Institute of Technology}\\
Harbin, China \\
mzsun@ir.hit.edu.cn}
\and
\IEEEauthorblockN{2\textsuperscript{nd} Sendong Zhao}
\IEEEauthorblockA{\textit{Faculty of Computing} \\
\textit{Harbin Institute of Technology}\\
Harbin, China\\
sdzhao@ir.hit.edu.cn
}
\and
\IEEEauthorblockN{3\textsuperscript{rd}Jianyu Chen}
\IEEEauthorblockA{\textit{Faculty of Computing} \\
\textit{Harbin Institute of Technology}\\
Harbin, China \\
hcwang@ir.hit.edu.cn
    }
\and
\IEEEauthorblockN{4\textsuperscript{rd}Haochun Wang}
\IEEEauthorblockA{\textit{Faculty of Computing} \\
\textit{Harbin Institute of Technology}\\
Harbin, China \\
hcwang@ir.hit.edu.cn
    }
\and
\IEEEauthorblockN{5\textsuperscript{th}Bing Qin}
\IEEEauthorblockA{\textit{Faculty of Computing} \\
\textit{Harbin Institute of Technology}\\
Harbin, China \\
qinb@ir.hit.edu.cn
    }
}

\maketitle

\begin{abstract}
Retrieval-augmented Generation (RAG) has demonstrated potential in enhancing medical question-answering systems through the integration of large language models (LLMs) with external medical literature. LLMs can retrieve relevant medical articles to generate more professional responses efficiently. However, current RAG applications still face problems. They generate incorrect information, such as hallucinations, and they fail to use external knowledge correctly. To solve these issues, we propose a new method named M-Eval. This method is inspired by the heterogeneity analysis approach used in Evidence-Based Medicine (EBM). Our approach can check for factual errors in RAG responses using evidence from multiple sources. First, we extract additional medical literature from external knowledge bases. Then, we retrieve the evidence documents generated by the RAG system. We use heterogeneity analysis to check whether the evidence supports different viewpoints in the response. In addition to verifying the accuracy of the response, we also assess the reliability of the evidence provided by the RAG system. Our method shows an improvement of up to 23.31\% accuracy across various LLMs. This work can help detect errors in current RAG-based medical systems. It also makes the applications of LLMs more reliable and reduces diagnostic errors.
\end{abstract}

\section{Introduction}


In medical research, artificial intelligence (AI) technologies have long been recognized for their potential to assist in clinical diagnosis and medical studies~\cite{holmes2004artificial}. With the rise of powerful medical large language models (LLMs), researchers are discovering that applying these models to various medical tasks can greatly improve efficiency~\cite{thirunavukarasu2023large, alberts2023large}. Recently, these models have achieved impressive results across multiple-choice-based medical exam datasets. In some medical question-answering tasks,  they have even been regarded as surpassing human medical experts~\cite{singhal2025toward}. As a result,  researchers are working with hospitals to improve these models and better integrate them into healthcare services.
\begin{figure}[t]
    \centering
    \includegraphics[width=1.0\linewidth]{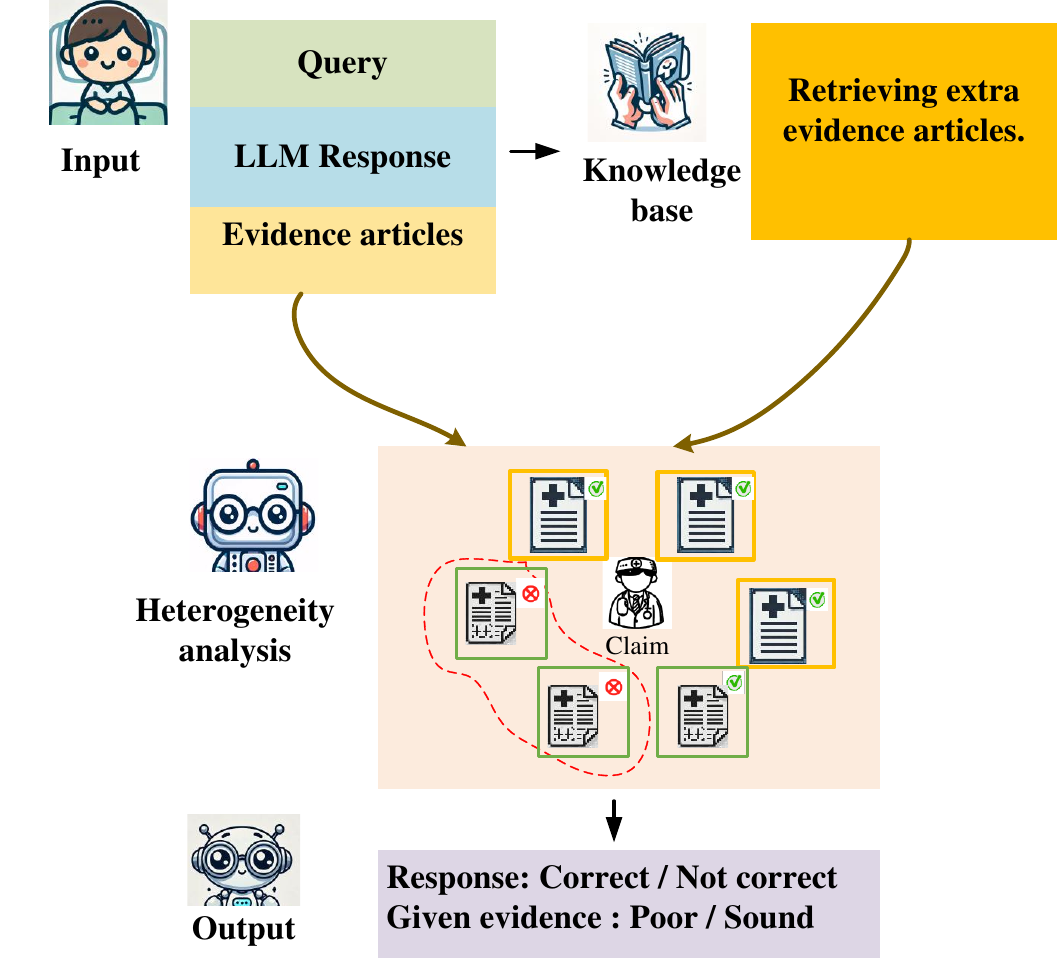}
    \caption{The task of the M-Eval checking system. M-Eval is designed to detect factual errors in the responses of medical RAG systems and analyze the quality of their evidence. The output of the task should include the label of the given responses and the evaluation of the evidence.  }
    \label{fig:1}
\end{figure}
\begin{figure*}[htp]
    \centering
    \includegraphics[width=1.0\linewidth]{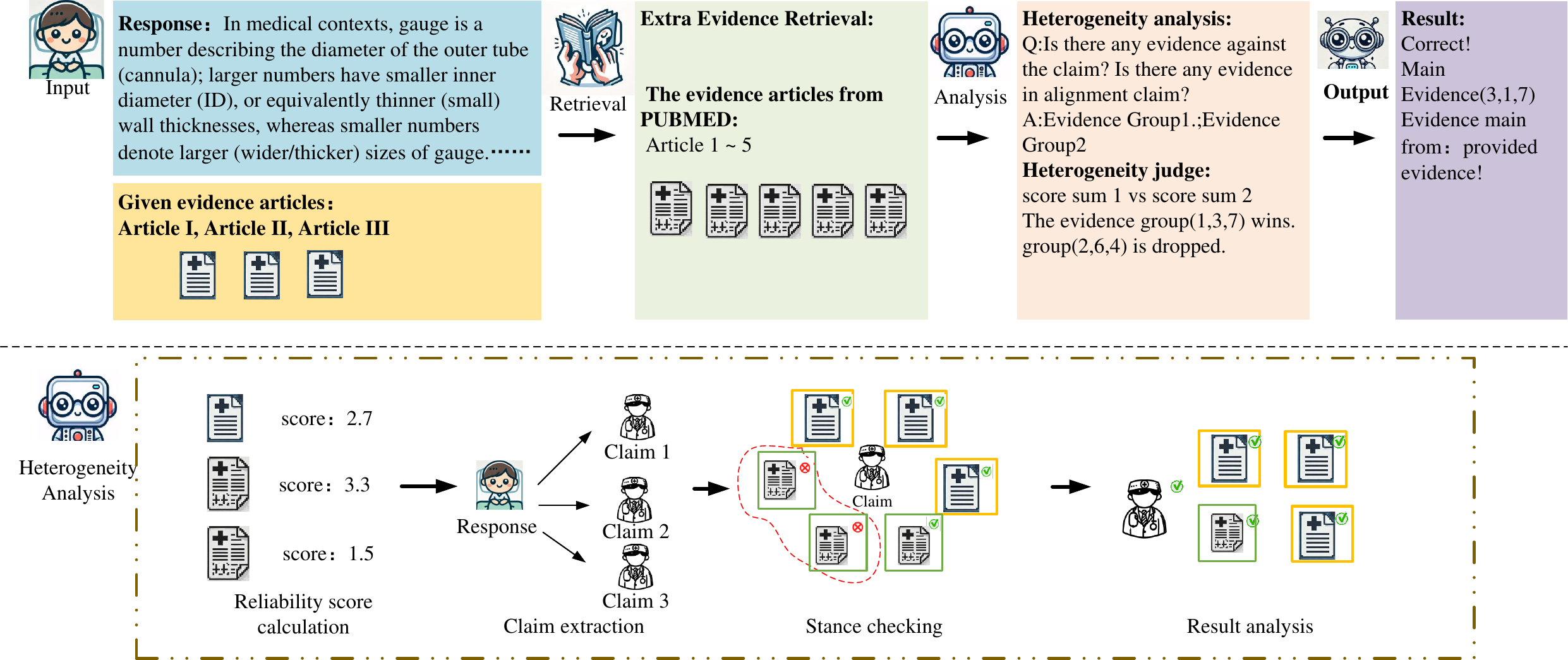}
    \caption{The pipeline of the M-Eval checking system. The main part of the system is the heterogeneity analysis detailed at the bottom side. We calculate the reliability score and the stance on the claim of extra evidence and given evidence. The reliability score is based on their revised date, publication type, and mesh heading. Then we test their stance on the claim and analyze the final label of the claim. }
    \label{fig:2}
\end{figure*}

However, the deployment of LLMs in healthcare faces challenges, mainly due to hallucinations, which are a known issue with these models~\cite{perkovic2024hallucinations,huang2024survey}. Hospital researchers argue that if the model cannot guarantee the accuracy of its outputs, no one can be held responsible for the consequences of any mistakes. To address this problem, researchers have proposed various methods to detect and reduce hallucinations in LLMs. One such approach is Retrieval-augmented Generation (RAG), which has proven effective~\cite{fan2024survey}. RAG works by retrieving evidence from medical literature databases, extracting relevant papers, and assisting the LLMs in generating their responses. This approach aligns closely with the principles of Evidence-Based Medicine (EBM). However, despite its effectiveness, RAG has limitations. Because medical literature is constantly evolving, the documents retrieved by the model often present differing viewpoints. The inconsistent articles retrieved may confuse the model and lead to errors in its responses. Additionally, since the model has its internal knowledge, it may occasionally generate responses that don't fully match the retrieved evidence.~\cite{gupta2024comprehensive}


To address these issues, we have developed a backend verification system for the RAG-based medical question-answering outputs. Our approach is inspired by the heterogeneity analysis used in EBM~\cite{dersimonian2015meta}. We use this analysis to evaluate all the claims by relevant evidence and verify the accuracy of the model's responses. As shown in Figure ~\ref{fig:1},  we first separate the output from the RAG system into two parts: the response and the evidence. By calculating the relevance of the questions, we gather the claims that need to be verified. Then we search for relevant articles in an additional PubMed database and combine them with the evidence given in the input. All the evidence articles will be checked if the viewpoints of the evidence align with the claim. Articles with differing opinions will be compared based on the reliability scores of their groups, and the final claim label will be decided accordingly.




We replicate the heterogeneity analysis from EBM and apply its principles to validate the outputs of RAG in the medical field. Our method effectively identifies errors in the responses and evaluates the quality of the evidence provided by the RAG system. It also alerts users about the relevance and freshness of the evidence. Our contributions are as follows:
\begin{itemize}
    \item  We propose a medical model output verification system based on EBM, which reduces the chances of users being misled by incorrect responses and evidence.
    \item  We evaluate the reliability and stance of the evidence, helping users assess the evidence extraction process in the given RAG answering system.
    \item  By simplifying the heterogeneity analysis in EBM, we introduce a multi-evidence medical response detection method, improving the model's experimental capability in such scenarios.
\end{itemize}
\section{Related works}
\subsection{Medical LLMs Application}

The advent of powerful LLMs like GPT-4 by OpenAI has drawn significant attention due to their impressive question-answering capabilities. As a result, LLMs have been introduced in various specialized fields, including the military, healthcare, and law \cite{thirunavukarasu2023large,xie2023factreranker,shah2023creation,clusmann2023future}. For instance, Google launched \textit{Med-PaLM2} \cite{singhal2023towards}, a highly fine-tuned version of \textit{PaLM2}, specifically aligned with extensive medical knowledge curated by experts and scholars. Google has claimed that \textit{Med-PaLM2} outperforms current medical professionals, which has generated considerable excitement in the research community. This development has spurred the creation of several other specialized models, such as \textit{ChatDoctor}~\cite{li2023chatdoctor}, \textit{MedAlpaca}~\cite{han2023medalpaca}, \textit{PMC-LLaMA}~\cite{wu2024pmc}, and \textit{BenTsao}~\cite{wang2023huatuo}. Experts in the respective languages typically evaluate these models, revealing a need for more efficient and effective evaluation methods.

With the advent of RAG technology, the medical field has a strong need for a generation approach that allows LLMs to reference external knowledge in a logical and evidence-based manner. Some multimodal researchers have adopted RAG to interpret and explain images or augment information~\cite{su2024implementing,zhu2024realm}. Meanwhile, another group of researchers has used methods similar to EBM to supplement knowledge for clinical questions, thus assisting models in generating responses. This approach enhances safety by making medical diagnoses more reasonable and helps address problems like language barriers or other factors~\cite{woo2024custom,xiong2024benchmarking}.

\subsection{Hallucination of LLMs on Medical Task}\label{2.2}

Considering several studies \cite{zhang2023siren, ji2023survey, umapathi2023med, manakul2023selfcheckgpt}, we conclude the classification and definition of the hallucination phenomenon in LLMs. Most researchers define hallucinations as the wrong outputs that conflict with real-world facts, fail to meet user requirements, or are unverifiable. In the medical domain,  model responses often diverge from traditional categories, resulting in phenomena such as excessive repetition or vague responses, and overly cautious answers. The meticulous nature of medicine and the inherent complexities of annotation have made current evaluation methods in medical tasks largely ineffective. For example, \textit{Med-PaLM2} \cite{singhal2023towards} relies on expert evaluations, while models like \textit{HuatuoGPT} ~\cite{zhang2023huatuogpt} perform self-checks using LLMs. Inspired by the ~\textit{Factools} methodology ~\cite{chern2023factool}, we try the traditional models to verify medical tasks. However, their performance decreases clearly when we change the input to the responses from another LLM. We also test the self-detection method using the model~\cite{ye2024boosting,zhang2024learning,miao2023selfcheck}. Compared to traditional tasks, the medical RAG task faces the challenge of ensuring evidence consistency, as medical literature is time-sensitive. As a result, the model is often misled when generating responses, and no matter how much introspection is applied, the model fails to understand the underlying issue.

\subsection{Hallucination of LLMs on Medical Task}\label{2.3}
In recent years, researchers have recognized and accepted EBM as an important discipline. EBM aims to make the best clinical decisions by integrating the best research evidence, clinical expertise, and patient preferences~\cite{subbiah2023next,mcmurray2021should}. With the advancement of information technology and data science, the application and development of EBM have been greatly enhanced. Traditional fact-checking work typically involves verifying claims against existing evidence, focusing on identifying conflicts between the claim and the evidence to detect erroneous facts~\cite{ye2024boosting,zhang2024learning,miao2023selfcheck}. This method is highly effective for conventional tasks. However, the outputs of medical LLMs are often highly misleading, which is a major reason they can easily misguide users. Traditional fact-checking methods using classical models struggle to perform such fine-grained error detection. Therefore, we extended fact-checking through heterogeneity analysis, resulting in the current MEVAL method.


\section{Method}
\subsection{Task Definition}
Our task is to validate the response of a typical RAG-based medical system and evaluate the credibility of the evidence it provides. The input format for this task is shown in the Figure \ref{fig:2}. We define a typical RAG medical system as consisting of at least two parts: the model’s response to the question and the evidence extracted during the generation of that response. We perform extra retrieval and fact-checking on this information and evaluate the claims in the responses based on heterogeneity analysis. Additionally, we assess the quality of the evidence provided by the RAG system. The factual evaluation of the results helps reduce the system’s diagnostic error rate, while the evidence quality evaluation aids in analyzing whether the front system suffers from outdated evidence or issues such as the model disregarding external evidence.

\subsection{Claim Extraction}
\begin{figure}
    \centering
    \includegraphics[width=1.0\linewidth]{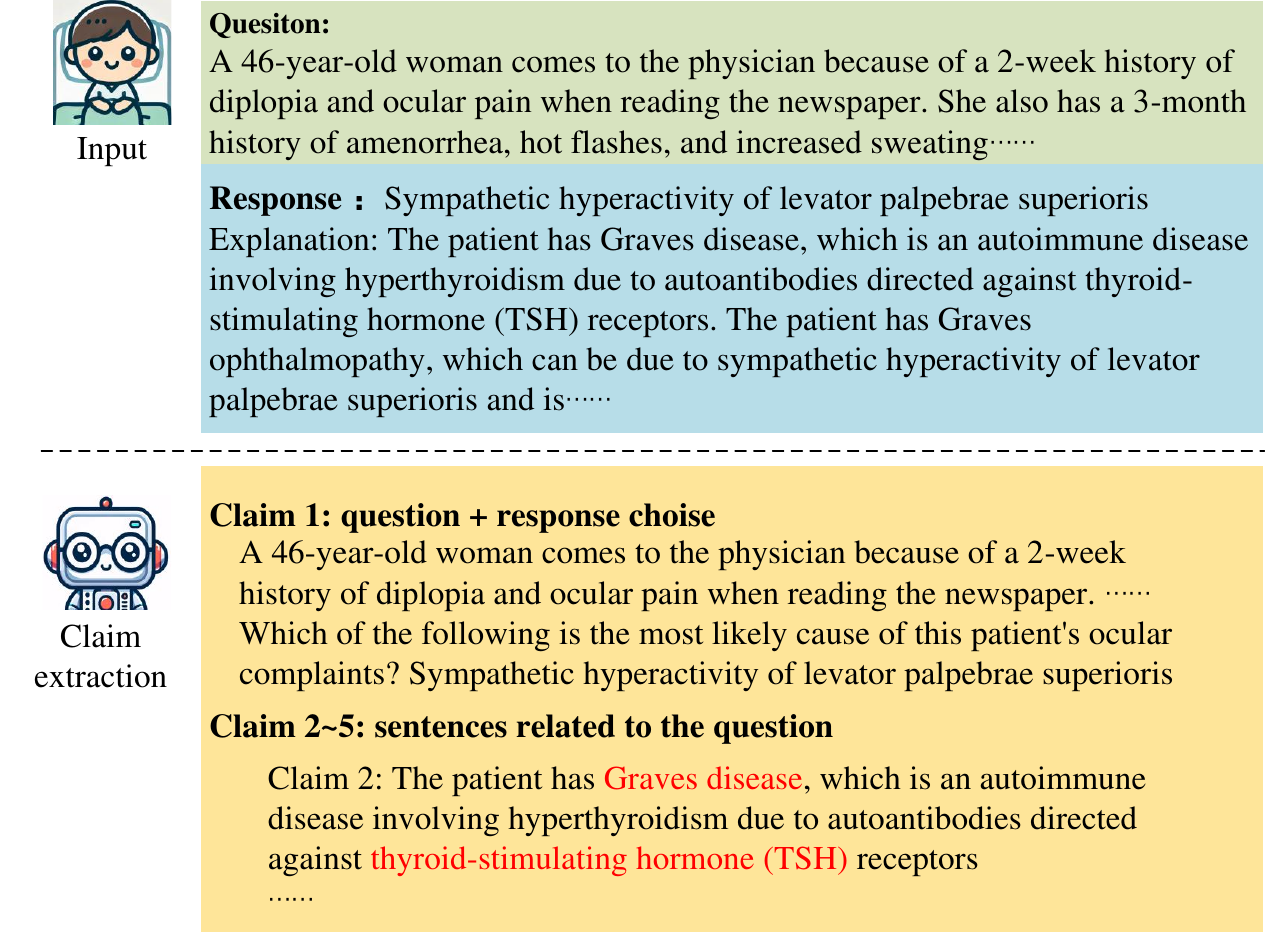}
    \caption{The example of the Claim extraction. our extraction is separated into two parts. The main claim is combined with the question and the choice in the response. The other claims are selected as the ones most related to the question.}
    \label{fig:3}
\end{figure}

Medical LLMs are often fine-tuned to improve safety or reduce user reliance, which leads to responses that often include a lot of irrelevant information. As shown in Figure ~\ref{fig:3}, the model may sometimes use avoidant phrases to discourage over-reliance on it. Additionally, to ensure the validity of the response, the model might generate lengthy explanations or repeat content from the evidence. These factors can complicate the evaluation of the responses. Therefore, the first step in assessing the RAG model's output is to extract the claims.

Our approach has two main parts. First, we use a medical-specific Spacy model to break down the RAG model's response into segments. Then, these segments are compared for similarity with the question using a model that has medical knowledge. We calculate the top four sentences that most contribute to answering the question and treat these sentences as the claims that could influence the model's performance. To prevent the model's response from relying too heavily on relevant literature, which could cause it to stray from the main question, we introduced an additional claim. This claim includes the question along with the model's chosen answer to reduce omissions. We denote these claims by $C=\{c_1, c_2, \dots, c_k\}$.

\begin{figure}[t]
    \centering
    \includegraphics[width=0.7\linewidth]{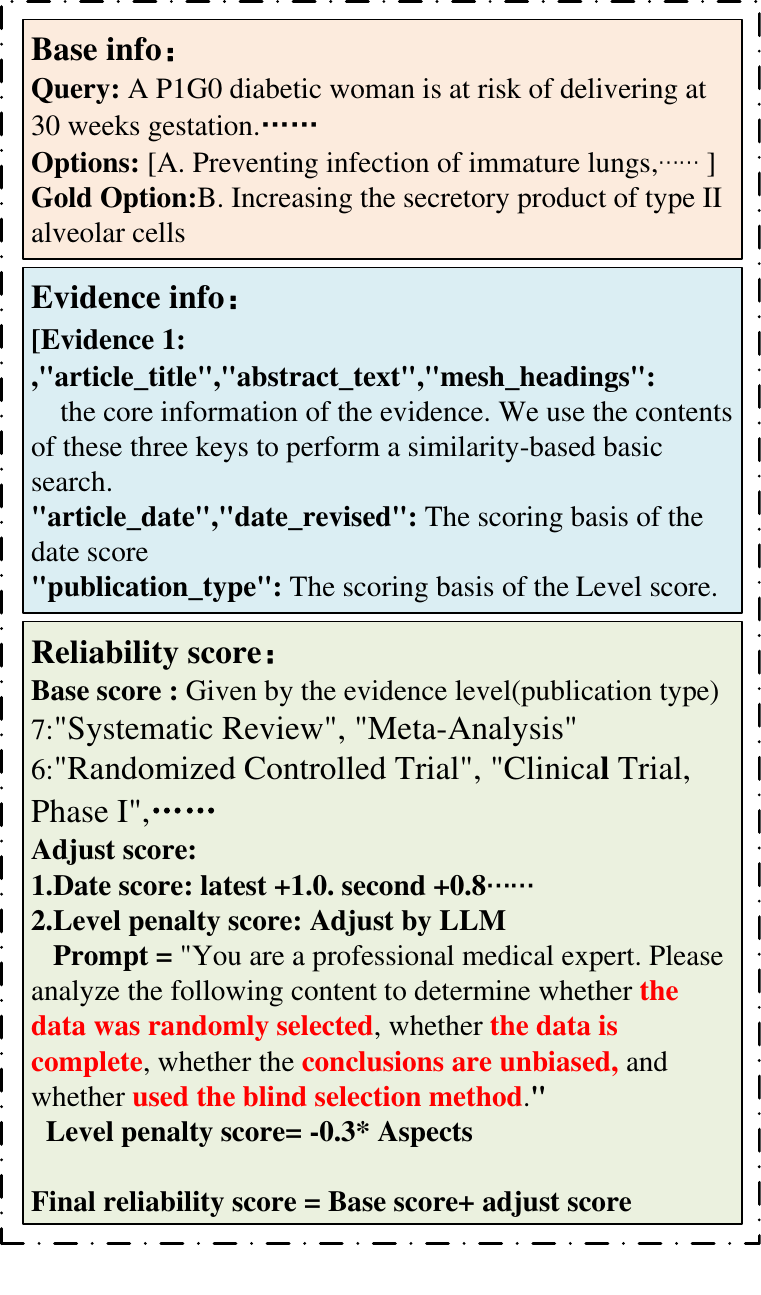}
    \caption{The method we calculate the reliability of each medicine article. For all the evidence, we need the information including publication date, publication type, and the mesh heading to analyze whether the article is reliable.  }
    \label{fig:4}
\end{figure}

\subsection{Evidence Retrieval}
After obtaining the claims to be verified, we need to search for relevant evidence. We use the PubMed dataset for this task. To ensure the reliability of the extracted evidence, we perform similarity calculations on the article's abstract, title, and MeSH terms simultaneously. The BM25 method is employed to extract the 15 most relevant articles. Then, we assess the reliability of each article based on its publication date and the journal in which it was published. As shown in the figure, we assign scores to each article following this rule-based approach and re-evaluate the evidence level of each article. Finally, we select the top $m$ highest-scoring articles as additional evidence for the entire experiment. We denote the given articles as $G=\{g_1, g_2, \dots, g_n\}$ and the extra retrieved articles as $E=\{e_1, e_2, \dots, e_m\}$. 

\begin{figure}[t]
    \centering
    \includegraphics[width=0.6\linewidth]{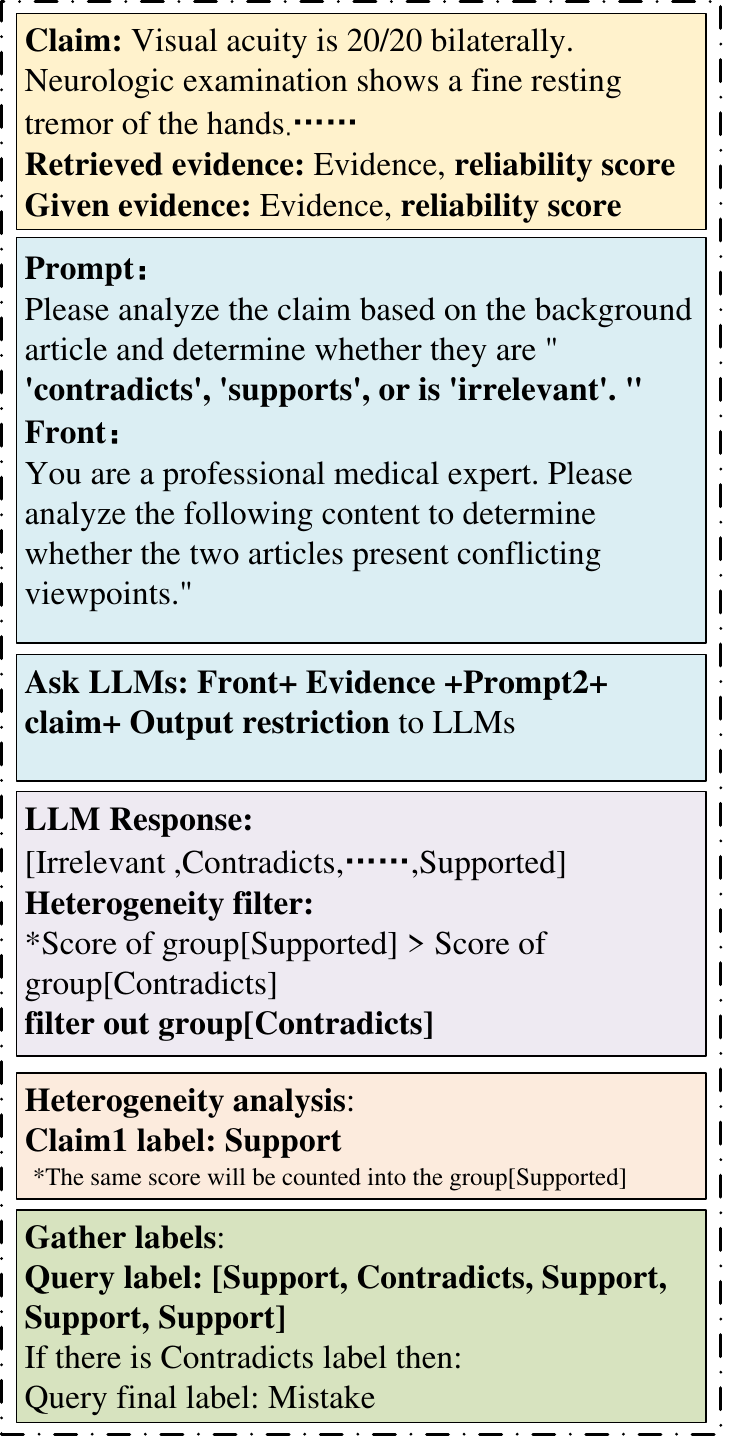}
    \caption{The detailed pipeline of the Heterogeneity analysis. We utilize different LLMs to compare whether the knowledge in the evidence article supports the claim. And we gather the claims label to get the final label of the response. }
    \label{fig:5}
\end{figure}
\vspace{-10pt}  

\subsection{Heterogeneity Analysis}
After identifying the claims to verify, we search for relevant evidence using the PubMed database~\cite{white2020pubmed}. To ensure the reliability of the extracted evidence, we calculate the similarity of the article’s abstract, title, and keywords. We use the BM25 method to extract the 15 most relevant articles. Then, we evaluate each article’s reliability based on its publication date and the journal in which it was published. As shown in Figure~\ref{fig:4}, we assign scores to each article using this rule-based approach and reassess the quality of the evidence. Finally, we select the top nine highest-scoring articles as additional evidence for the experiment. We denote the reliability score of each article as ${R}(e_i)$ and also the same for the ${R}(g_n)$

\begin{equation}
{y_m}(c_k, e_i) = \begin{cases}
1, & \text{if $e_i$ supports $c_k$} \\
-1, & \text{if $e_i$ contradicts $c_k$} \\
0, & \text{else}
\end{cases}
\end{equation}

  
Next, we calculate the scores for all evidence related to the claim. We refer to the defination in the DerSimonian–Laird random‐effects model in Formula 2 and Formula 3~\cite{dersimonian2015meta}.

\begin{equation}
Q
\;=\;
\sum_{i=1}^kq_i
\;=\;
\sum_{i=1}^k
w_i\,\bigl(y_i - \frac{\sum_{i=1}^k w_i\,y_i}
     {\sum_{i=1}^k w_i})^2\quad
\end{equation}

\begin{equation}
\tau^2_{\mathrm{DL}}
\;=\;
\max\!\Biggl\{
\frac{Q - (k - 1)}
     { \sum_{i=1}^k w_i \;-\;\frac{\sum_{i=1}^k w_i^2}{\sum_{i=1}^k w_i}}
\;,\;0
\Biggr\}
\end{equation}
This model requires the random‐effects variance $\tau^2_{DL}$ and Cochran’s $Q$ to show the heterogeneity of the evidence. Higher values of them indicate greater heterogeneity in the evidence.
To ensure consistency, we filter studies to minimize both $\tau^2_{DL}$ and $Q$. The formula requires the evidence labels $y_i$ and each study’s sampling variance $v_m={w^{-1}_i}$. However, Most studies do not report sampling variance $v_i$ in their abstracts or metadata. We have to set the $v_i$ as a constant. Also,
We take the reliability score to give higher weight to more authoritative and reliable evidence. 
Therefore, we define the heterogeneity score in Formula 4 to define the label of the remaining evidence. Finally, if all five claims are correct, the response is deemed accurate. If any claim is incorrect, the response is classified as having factual errors.
\begin{align}
\small
    \mathcal{M}_e(c) = & \sum_{j=1}^n 
    q\big(c , g_{j}\big) r\big(g_{j})\nonumber \\
     +& \sum_{i=1}^m 
    q\big(c , e_{i}\big) r\big(e_{i})
\end{align}



\section{Experiments and Results}
\subsection{Experiments Construction}



\begin{figure}[h]
    \centering
    \includegraphics[width=0.8\linewidth]{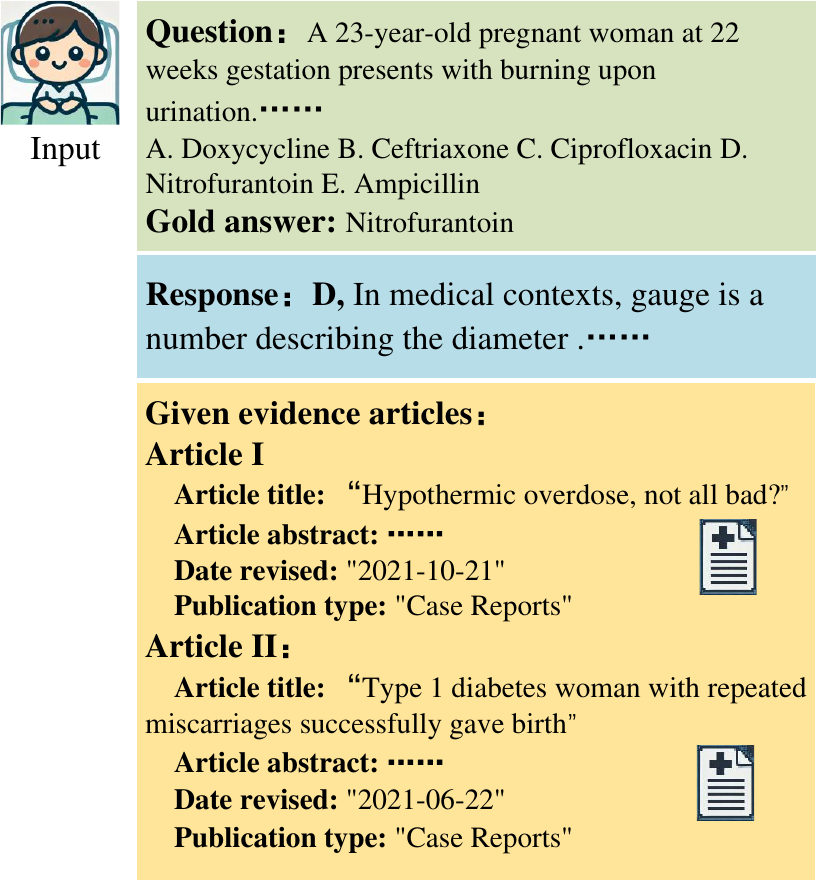}
    \caption{The Method to construct the datasets used in this work. The Finer group is given better knowledge evidence and the accuracy is higher than the normal group.  }
    \label{fig:input}
\end{figure}

\textbf{Datasets} The input format is shown in Figure~\ref{fig:input}. This task consists of a response from an LLM, along with two pieces of valid evidence. The information for these pieces of evidence should be listed separately, and the articles should provide detailed and relevant information to ensure the authenticity and validity of the evidence. To test the effectiveness of our method, we create two evidence groups to generate responses with different levels of response and evidence quality. As shown in the figure, we extract up to 15 pieces of evidence for each question. First, we rank these pieces based on their relevance to the model. Using relevance scores and the evidence level of each article, we filter and reorder the evidence. The top three most relevant pieces are selected as the Finer group evidence, and the remaining 12 are randomly selected from the 15 to form the Random group evidence. We then provide both evidence groups to the LLMs for response generation. This results in 2000 responses with an accuracy rate of 54\% and 2000 responses with an accuracy rate of 24\%. However, we observe that when the model’s response accuracy is too low, the outputs contain a substantial amount of noise. Some LLMs can not realize the claim, which leads to meaningless responses. Therefore, we prepare another pair of datasets. These response have 150 responses with an accuracy rate of 43.3\% from the Finer group and 147 responses with an accuracy rate of 31.4\% from the Random group. The second dataset is labeled * in Table~\ref {tab:1} and Table~\ref{tab:2}.

\textbf{Evaluation} After determining the correctness of the given response, we also evaluate the evidence provided by the front-end RAG model. During the heterogeneity analysis, we label each piece of evidence based on whether it supports or opposes the final outcome. We track each piece of evidence’s contribution to the voting process. If the evidence consistently supports the side opposite to the final outcome, it misguides the model’s response. If the evidence casts many irrelevant votes, it suggests a problem with the front-end retrieval module. On the other hand, if the evidence consistently aligns with the final result, it is considered strong and appropriate. After evaluating the contribution of the three types of internal evidence, we can assess the overall effectiveness of the front-end RAG model’s evidence.

\subsection{Main Result}
\begin{table}[t]

    \centering
    \caption{Accuracy(\%), Recall(\%), and Specificity(\%) of M-Eval and other baselines.  }
    \renewcommand{\arraystretch}{1.0} 
    \begin{tabular}{cp{1.0cm}|p{0.55cm}p{0.55cm}p{0.55cm}|p{0.55cm}p{0.55cm}p{0.55cm}}
        \toprule
        \toprule
        \multicolumn{2}{c|}{\multirow{2}{*}{\textbf{Method}}} & \multicolumn{3}{c|}{\textbf{Random Group}} & \multicolumn{3}{c}{\textbf{Finer Group}} \\
        & & Acc& Rec & Spe  & Acc & Rec & Spe   \\

        \midrule
        \multirow{3}{*}{Qwen2.5-14B} & M-Eval & 72.10 & 9.91 & 90.89 & 72.1 & 10.09  & 90.94\\
         & w/o Evi & 68.70 & 25.54 & 81.26 & 62.7 & 68.13 & 55.74 \\
        & Self & 75.55 & 3.00 & 97.52  & 46.55& 2.84& 99.11 \\
        \midrule
        \multirow{3}{*}{Mistral-7B} & M-Eval & 71.45 & 10.78 & 89.78 & 71.30 & 10.52  & 89.77\\
         & w/o Evi & 32.22& 89.27 & 12.28 & 54.60 & 98.90 & 0.26 \\
        & Self & 70.40 & 15.67 & 86.63  & 49.65& 25.46& 78.55 \\
        \midrule
        \multirow{3}{*}{Gemma2-9B} & M-Eval & 62.95 & 23.92 & 74.74 & 63.20& 24.25  & 75.03\\
         & w/o Evi  & 45.15 & 61.69 & 39.97 & 62.85 & 68.13 & 56.57\\
         & Self & 23.84 & -- & -- & 54.83 & --  & --\\
         \midrule
         \multirow{3}{*}{Gemma1.1-7B*} & M-Eval & 59.86& 26.53 & 76.53  & 60.67 & 25.00 & 75.47 \\
         & w/o Evi  & 54.90 &55.10 & 39.80 & 55.00 &43.18 & 61.32  \\
         & Self & 58.90 & 24.49 & 80.61 & 57.33 & 31.82 & 67.92  \\
          \midrule
        \multirow{3}{*}{Llama3-8B*} & M-Eval & 60.78  & 50.00 & 65.71 & 53.49&46.67 & 57.14 \\
         & w/o Evi & 35.37 & 91.84 & 7.14 &49.33 & 87.69 & 20.00  \\
         & Self & 37.41 & 83.67 &14.29 & 44.00 & 72.31 & 22.35 \\
         \midrule
        \multirow{3}{*}{Qwen2.5-7B*} & M-Eval &60.54  & 16.33 & 82.65 & 58.67 &  15.91  & 76.42\\
         & w/o Evi  & 56.67 & 33.33 & 66.67 & 50.00 & 31.47 & 68.52 \\
        & Self & 54.90 & 55.10 & 39.80 & 46.67  & 9.09& 90.57 \\
        \bottomrule
         \bottomrule
    \end{tabular}

    \label{tab:1}
\end{table}
\begin{table*}[t]
    \centering
    \caption{The details of M-Eval accuracy(\%) with different numbers of extra articles. }
    \label{tab:2}
    \renewcommand{\arraystretch}{1.25} 
    \begin{tabular}{cp{1.2cm}|p{0.6cm}p{0.6cm}p{0.6cm}p{0.6cm}p{0.6cm}|p{0.6cm}p{0.6cm}p{0.6cm}p{0.6cm}p{0.6cm}}
        \toprule
        \toprule
        \multicolumn{2}{c|}{\multirow{2}{*}{\textbf{Method}}} & \multicolumn{5}{c}{\textbf{Random Group}} & \multicolumn{5}{c}{\textbf{Finer Group}}  \\
        & & 1 & 2 & 3 & 4 & 5  & 1 & 2 & 3 & 4 & 5 \\
        \midrule
        \multirow{3}{*}{Gemma2-9B} & Acc & 62.95 & 61.15 & 59.75 & 58.40 & 57.10 & 63.20 & 61.50 & 60.35 & 59.65 & 58.00 \\
         & Rec  & 23.92 & 26.72 & 30.82 & 34.48 & 36.21 & 24.25 & 27.47 & 31.97 & 36.70 & 37.34\\
         & Spe  & 74.74 & 71.55 & 68.49 & 65.62 & 63.41 & 75.03 & 71.84 & 68.97 & 66.62 & 64.28 \\
        \midrule
        \multirow{3}{*}{Mistral-7B} & Acc  & 71.45 & 70.80 & 69.85 & 69.45 & 68.70 & 71.30 & 70.75 & 69.85 & 69.45 & 68.80 \\
         & Rec  & 10.78 & 12.07 & 14.01 & 16.38 & 16.59 & 10.52 & 12.45 & 14.38 & 17.17 & 17.38\\
         & Spe  & 89.78 & 88.54 & 86.72 & 85.48 & 84.38 & 89.77 & 88.46 & 86.70 & 85.33 & 84.42 \\
        \midrule
        \multirow{3}{*}{Qwen2.5-14B} & Acc  & 72.30 & 71.75 & 71.00 & 70.65 & 70.05 & 72.25 & 71.90 & 71.20 & 70.85 & 70.25 \\
         & Rec  & 9.91 & 11.64 & 13.79 & 14.87 & 15.95 & 39.48 & 46.36 & 54.08 & 59.24 & 62.68\\
         & Spe  & 90.89 & 89.45 & 87.7 & 86.85 & 85.74 & 90.94 & 89.77 & 88.14 & 87.16 & 86.18 \\
         \midrule
        \multirow{3}{*}{Llama3-8B*} & Acc  & 60.78 & 58.82 & 62.75&62.75&60.78& 58.14 &60.47 &48.84 &51.16&53.49\\
         & Rec  & 37.50 &25.00 & 37.50& 56.25 &50.00 & 66.67 & 66.67& 53.33 & 46.67 &46.67\\
         & Spe  & 71.43 & 74.29& 74.29& 65.71 &65.71  & 53.57 & 57.14& 46.43& 53.57 &57.14 \\
         \midrule
        \multirow{3}{*}{Qwen2.5-7B*} & Acc  &55.78&59.86&60.54&60.54&60.54& 54.00 & 54.67 & 55.33&57.33&58.67\\
         & Rec  & 22.45 & 22.45& 20.41& 18.37 &16.33  & 27.27 & 18.18& 18.18& 18.18 &15.91\\
         & Spe  & 72.45& 78.57& 80.61& 81.63 &82.65  & 65.09 & 69.81& 70.75& 73.58 &76.42 \\
         \midrule
        \multirow{3}{*}{Gemma1.1-7B*} & Acc  &  59.18& 61.22&59.86&59.86&59.86&57.33&58.67&58.67&59.33&60.67\\
         & Rec  & 24.49 & 28.57& 26.53& 26.53 &26.53 & 29.55 &27.27& 25.00& 22.73 &25.00\\
         & Spe  & 76.53 & 77.55& 76.53& 76.53 &76.53  & 68.87 & 71.70& 72.64& 74.53 &75.47 \\

        \bottomrule
         \bottomrule
    \end{tabular}

\end{table*}

In Table~\ref {tab:1} and Table~\ref{tab:2}, we show the performance improvement of M-Eval compared to traditional baselines. As additional evidence is progressively added, the model's final accuracy gradually stabilizes, and both its Recall (Rec) and Specificity (Spe) steadily adjust, indicating that the method's judgments are becoming more balanced. The numbers 1, 2, 3, 4, and 5 under the dataset name represent the number of extra evidence articles provided. Our scenario involves judging misdiagnoses in a zero-shot environment based on the outputs of LLMs. 

We select two baselines for comparison in our experiment. Only LLMs capable of handling different types of noisy information while still producing correct outputs were considered. In \textbf{w/o Evi}, the LLM uses no evidence and only relies on its internal knowledge to evaluate the given query. In \textbf{Self}, we instruct the model to follow the evidence provided with the RAG response, trusting all the articles given by the previous RAG system. However, some LLMs exhibited a pronounced bias when executing the baseline method. Notably, the Gemma2-9B model classifies all outcomes as correct. As shown in Table~\ref {tab:1}, our M-Eval method demonstrates a significant improvement over traditional direct evaluation with LLMs. This improvement is consistent, regardless of the quality of evidence or the actual accuracy ratio in the dataset.

\subsection{Extra Evidence Analysis}
\begin{figure*}[h]
    \centering
    \includegraphics[width=1.0\linewidth]{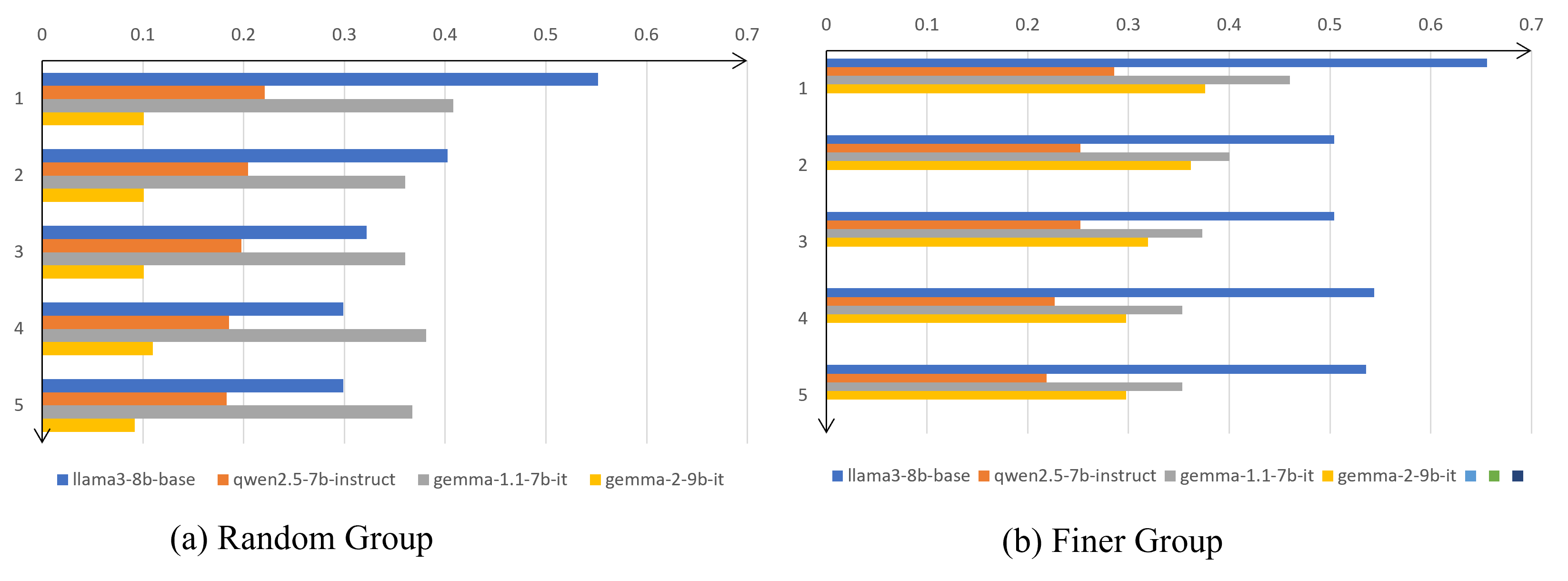}
    \caption{The result of the proportion of contribution evidence queries. We assess whether the given evidence aligns with the overall results of all the evidence to determine its contribution to the outcome. The x-axis represents the proportion of contributions made by the given evidence, and the y-axis represents the number of additional extracted articles.
}
    \label{fig:7}
\end{figure*}

We also evaluate the quality of the provided evidence. As shown in Table~\ref{tab:2}, we show the performance changes by different numbers of extra evidence. We find that the M-Eval method consistently performs well across datasets exhibiting diverse distributions. Also, we investigate how additional evidence influences the final judgment. The lines in Figure~\ref{fig:7} represent the proportion of queries whose given evidence matches the final determined label. We observe that when a small amount of evidence is initially provided, almost all final labels are consistent with the initial labels. However, as the amount of additional evidence increased, this proportion noticeably decreases. The final stable value depends on the quality of the provided evidence. This also indicates that the given evidence contains significant gaps, and the original evidence often misguides the evaluation of the model's output.

\subsection{Ablation Study}
To validate the effectiveness of each component of our method, we conduct ablation experiments. Specifically, we test the impact of the number of evidence pieces in various parts of the model. First, in the ablation experiment on reliability score validation, we check the accuracy by randomly assigning scores to each piece of evidence and observing how this affects the final result’s accuracy. In the experiment on heterogeneity analysis, we examine how the results would change if any claim contained an error. Finally, we validate our additional evidence extraction component by removing all the extra evidence and observing the changes in the results, as shown in Table~\ref{tab:3}.
\begin{table}[h]
    \centering
    \small
    \caption{The ablation studies of M-Eval.  }
    \renewcommand{\arraystretch}{1.0} 
    \begin{tabular}{cp{1.2cm}|p{0.7cm}p{0.7cm}p{0.7cm}}
        \toprule
        \toprule
        \multicolumn{2}{c|}{\multirow{2}{*}{\textbf{Method}}} & \multicolumn{3}{c}{\textbf{Llama3}}  \\
        &  & Acc& Rec & Sep   \\
        \midrule
        \multirow{3}{*}{Random} & M-Eval & 62.75 &65.71& 56.25 \\
         & A-Reli. & 60.78 & 50.00 & 65.71\\
         & A-Hete. & 59.79 & 16.67 & 79.10  \\
        & A-Retr. & 52.17 & 54.55 & 51.06 \\
         \midrule
        \multirow{4}{*}{Finer} & M-Eval & 53.49 & 46.67  & 53.57 \\
         & \textit{A-Reli}.  & 48.39 & 41.67 & 52.63 \\
        & \textit{A-Hete}. & 45.16 & 20.83 & 60.53  \\
        & \textit{A-Retr}.& 43.55 & 70.83 & 26.32 \\
        \bottomrule
         \bottomrule
    \end{tabular}

    \label{tab:3}
\end{table}

\textbf{A-Reli}: In the first ablation experiment, we randomly assign a reliability score between 0 and 7 to each article using a random seed. As shown in Table~\ref{tab:3}, we observe a significant drop in model performance. When the score is randomly set to a specific value, accuracy approaches the scenario where all responses are judged incorrect, which increases accuracy. In this case, Recall (Rec) drops significantly, while Specificity (Sep) rise to nearly 90\%. However, such results are not practical for real-world datasets, so we do not consider them valid.

\textbf{A-Hete}: In the second ablation experiment, we test how removing the heterogeneity analysis step would affect our conclusions. Instead of summing scores for each claim label, we simply check if there is any evidence that contradicts the claim. If any piece of evidence negates the claim, we consider it incorrect. This approach improves accuracy for datasets with many negative examples. However, as shown in Table~\ref{tab:3}, rejecting many positive labels causes the overall accuracy to decrease. This result highlights the importance of properly analyzing and screening the evidence.

\textbf{A-Retr}: In the third ablation experiment, we test the system's performance when no additional evidence is used, and only the provided evidence is considered. We find that while our method still detects issues effectively, the accuracy is much lower compared to the main experiment, where high-quality evidence is used.

\section{Conclusion}
LLMs have gained significant attention for their strong comprehension and generation capabilities. In the medical field, there is a growing need for such powerful automated tools to assist with medical tasks. However, even with RAG, LLMs still generate factual errors. To address this, we propose a method to evaluate the responses and evidence of RAG. Our method shows significant improvements in experiments compared to self-checking approaches. We hope that our work will help doctors identify factual errors and make them more practical for use in the medical field.

\section*{Limitations of the work}

We are unable to fully replicate every detail of the meticulous process involved in meta-analysis. To gain insights into the principles of meta-analysis, we consult several medical experts. We find that the heterogeneity analysis step in meta-analysis typically requires obtaining detailed experimental data from each medical publication for calculation and comparison. However, our LLM cannot contact the authors of each study to request this data. Therefore, our heterogeneity analysis is more of an approximation, imitating the original heterogeneity analysis process. We have experimented with data-driven methods based on native approaches, but this process is constrained by the dataset we had.


\printbibliography

\end{document}